\documentclass[letterpaper]{article} 
\usepackage{aaai25}  
\usepackage{times}  
\usepackage{helvet}  
\usepackage{courier}  
\usepackage[hyphens]{url}  
\usepackage{graphicx} 
\usepackage{natbib}  
\usepackage{caption} 
\usepackage{algorithm}
\usepackage{algorithmic}

\usepackage{booktabs}

\usepackage{newfloat}
\usepackage{listings}
\DeclareCaptionStyle{ruled}{labelfont=normalfont,labelsep=colon,strut=off} 
\floatstyle{ruled}
\newfloat{listing}{tb}{lst}{}
\floatname{listing}{Listing}
\title{Is Your Image a Good Storyteller?}
\author{
    Xiujie Song\textsuperscript{\rm 1}, 
    Xiaoyi Pang\textsuperscript{\rm 1},
    Haifeng Tang\textsuperscript{\rm 2},
    Mengyue Wu\textsuperscript{\rm 1}\thanks{Corresponding authors.}, 
    Kenny Q. Zhu\textsuperscript{\rm 3}\footnotemark[1] 
}
\affiliations{
    \textsuperscript{\rm 1} X-LANCE Lab, Department of Computer Science and Engineering \\
    MoE Key Lab of Artificial Intelligence, AI Institute\\
    Shanghai Jiao Tong University, Shanghai, China \\
    \textsuperscript{\rm 2}China Merchants Bank Credit Card Center, Shanghai, China  \\
     \textsuperscript{\rm 3}University of Texas at Arlington, Arlington, Texas, USA \\

    \textsuperscript{\rm 1} \{xiujiesong,  fointpang, mengyuewu\}@sjtu.edu.cn,
    \textsuperscript{\rm 2} thfeng@cmbchina.com,
    \textsuperscript{\rm 3} kenny.zhu@uta.edu

}

\usepackage{bibentry}

\begin{document}

\maketitle

\begin{abstract}
    Quantifying image complexity at the entity level is straightforward, but the assessment of semantic complexity has been largely overlooked.
    In fact, there are differences in semantic complexity across images.
    Images with richer semantics can tell vivid and engaging stories and offer a wide range of application scenarios.
    For example, the Cookie Theft picture is such a kind of image and is widely used to assess human language and cognitive abilities due to its higher semantic complexity. 
    Additionally, semantically rich images can benefit the development of vision models, as images with limited semantics are becoming less challenging for them. 
    However, such images are scarce, highlighting the need for a greater number of them.
    For instance, there is a need for more images like Cookie Theft to cater to people from different cultural backgrounds and eras.
    Assessing semantic complexity requires human experts and empirical evidence. 
    Automatic evaluation of how semantically rich an image will be the first step of mining or generating more images with rich semantics, and benefit human cognitive assessment, Artificial Intelligence, and various other applications.
    In response, we propose the Image Semantic Assessment (ISA) task to address this problem. 
    We introduce the first ISA dataset and a novel method that leverages language to solve this vision problem. 
    Experiments on our dataset demonstrate the effectiveness of our approach.
\end{abstract}


\begin{links} 
\link{Data and code}{https://github.com/xiujiesong/ISA} 
\end{links}


\section{Introduction}

\begin{figure}[htbp]
    \centering
    \includegraphics[scale=0.36]{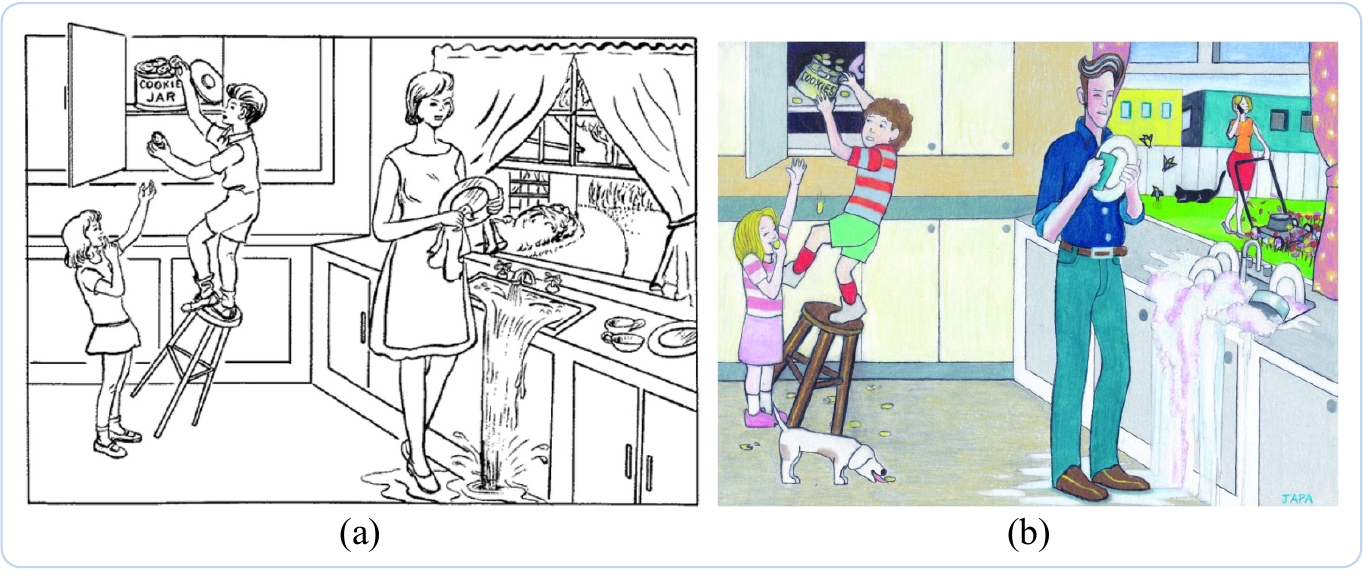}
    \caption{The Cookie Theft pictures. (a) is the original version and (b) is an updated version.}
    \label{ct}
\end{figure}

How complex can a picture be?
What kind of story can be told via a single picture? 
As the saying goes, ``a picture is worth a thousand words''. 
However, not every picture contains such rich information.
The Cookie Theft picture (Figure~\ref{ct} (a)) is a good exemplar of complex semantic information expressed via visual language. 
It is a well-known picture commonly used to assess language and cognitive abilities in humans. 
It was first introduced in the Boston Diagnostic Aphasia Examination published in 1972~\cite{goodglass2001-ej} and remains widely utilized to this day.

Many studies~\cite{cummings2019describing, tasnim-etal-2022-depac} have revealed the reasons behind the success of this picture.
Its essence is being a ``good storyteller,'' capable of telling a complete and engaging story. 
Based on the research of the psychologists, two of its most important characteristics can be summarized as follows: 
(1) It contains a rich but not excessive number of entities, making it well-suited for eliciting longer narrative descriptions. 
(2) It is rich in semantics, enabling it to tell an interesting story.
The semantics are derived from reasonings made by observing the entities and their relationships in the image.
For instance, the Cookie Theft tells a story about two children attempting to steal cookies from a jar when their mother is not looking. 
The mother-child relationship between the characters in the image is deduced through further reasoning based on observing the content in the image.

Though the Cookie Theft picture is widely used, there are still limitations. 
It is outdated since it has been proposed for half a century and it cannot be well applied to different cultures~\cite{berube2019stealing, rethinkingct}.
To avoid these issues, people often have to modify or replace the image in different application scenarios~\cite{berube2019stealing, HUSSEIN2015152, DOMINGUEZ2006476, Oh2012ValidityAR, Prasad2012ValidationOT}.
For instance, Figure~\ref{ct} (b) is an updated version of the Cookie Theft.
This means more images of this kind are necessary. 

Besides, this kind of image is not only useful for humans but also for Artificial Intelligence (AI).
With the development of vision models, especially Large Vision-Language Models (LVLMs), their abilities are increasing rapidly. 
Simple images with less semantics are not challenging enough for them to understand or generate anymore, so more images with rich semantics will definitely be beneficial for both training and evaluation~\cite{song2024cognitive}.

The internet or existing image datasets contain a lot of images, including some high-quality images that we expect. 
However, due to their scarcity, identifying and locating these high-quality images amidst the vast array of webly images can be a daunting task. 
Therefore, efficient methods for scoring and selecting these images are crucial.
Furthermore, with the advancement of image generation models~\cite{Rombach2021HighResolutionIS, pmlr-v139-ramesh21a, Ramesh2022HierarchicalTI}, they are also increasingly capable of helping us generate more images.
Thus, automatic semantic complexity assessment can also be used to assess the semantic complexity of generated images.
Generally, it is the necessary path for obtaining semantically rich images.

Currently, though there are some research works about Image Assessment, like Image Quality Assessment (IQA)~\cite{Fang_2020_CVPR, ying2020patches}, Image Aesthetics Assessment (IAA)~\cite{ijcai2022p132, Yi_2023_CVPR}, and Image Complexity Assessment (ICA)~\cite{saraee2020visual, ic9600}, no one focuses on assessing the semantic complexity of images.
In order to fill this research blank, we propose the \textbf{Image Semantic Assessment (ISA)} task to assess the semantic complexity of images.

Considering entities are the foundation of semantics and the complexity requirements for these two aspects may vary in different application scenarios, ISA task assesses images from both two levels: 
1) At the \textbf{\textit{entity}} level, we assess the entity richness of images, similar to the idea of ICA task~\cite{ic9600}, which we refer to as the Entity Complexity Scoring task; 
2) At the \textbf{\textit{semantic}} level, we propose the Semantic Complexity Scoring task to assess the higher-level semantic complexity of images. Note that this sub-task is the core of our proposed ISA task.

To promote the research on ISA task, we built the first ISA dataset with 2,946 images.
Each image is annotated with the two corresponding scores by three annotators.
Besides, a corresponding method called \textbf{V}ision-\textbf{L}anguage collaborative \textbf{ISA} method (\textbf{VLISA}) is proposed for this novel task. 
It first uses a Large Vision-Language Model (LVLM), such as GPT-4o~\cite{gpt4}, as a feature extractor to extract semantic information in natural language form from images. 
Then, a regression model is trained to predict the score of images.
Our contributions are as follows:

1. As far as we know, we are the first to propose the ISA task, which aims to automatically assess semantic complexity in an image. 
It can be used to identify high-quality images with rich semantics and evaluate image generation models, etc. 

2. We construct the first ISA dataset, consisting of 2,946 images and human scores, that supports the ISA task. Our dataset includes images of varying semantic complexity, which helps models learn the ability to assess semantic complexity.

3. To effectively assess the semantic complexity of images, we propose a simple yet effective method that collaboratively utilizes language and visual information.
Experiments show that ISA task is challenging for traditional vision models like ViT~\cite{vit} and our proposed method significantly outperforms other baseline models on the Semantic Complexity Scoring task.

\section{ISA Dataset Construction}

In this section, we introduce our ISA data collection and annotation process, as well as the related data analysis.

\subsection{Data Collection}
We collected our images from Pinterest\footnote{https://www.pinterest.com/}. 
After collecting images, we filtered out duplicated images using imagededup\footnote{https://github.com/idealo/imagededup}. 
To ensure high quality, we also manually excluded low-quality images that were blurry, watermarked or contained unnecessary text. 
After filtering, we finally retained 2,946 images in our dataset.

\begin{figure}[h]
    \centering
    \includegraphics[scale=0.48]{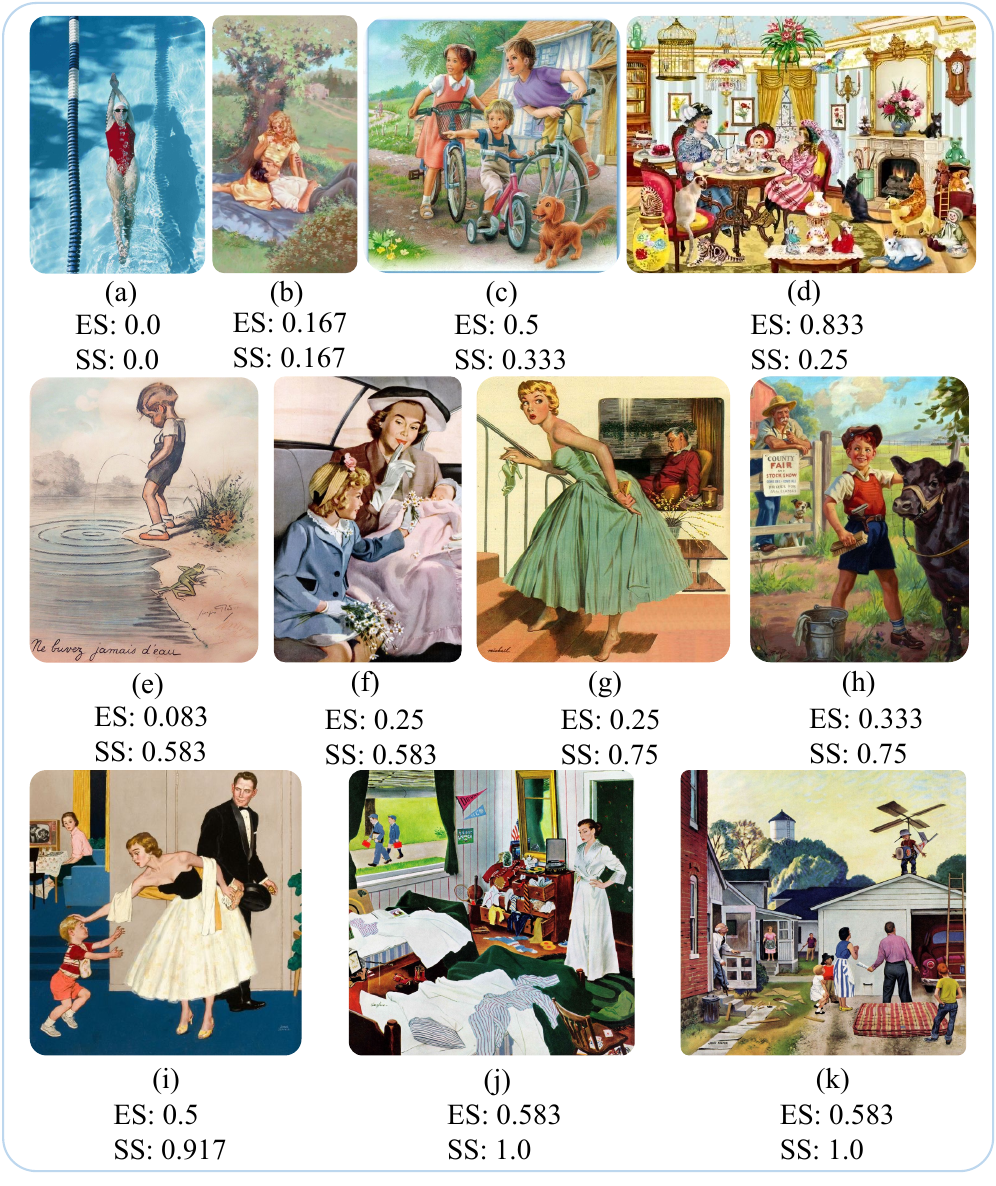}
    \caption{Samples from the proposed ISA dataset. ES and SS stand for Entity Score and Semantic Score respectively.}
    \label{samples}
\end{figure}

\subsection{Data Annotation}

For each image, we annotate it with two scores: an Entity Score and a Semantic Score. 
They correspond to the Entity Complexity Scoring task and the Semantic Complexity Scoring task, respectively.
For each score, the images are first annotated on a scale from 1 (Low) to 5 (High).
Then, these scores are normalized to the [0,1] range~\cite{ic9600}, and the average of these normalized scores is calculated as the final score.

\subsubsection{Entity Score}
\begin{figure}[ht]
    \centering
    \includegraphics[scale=0.48]{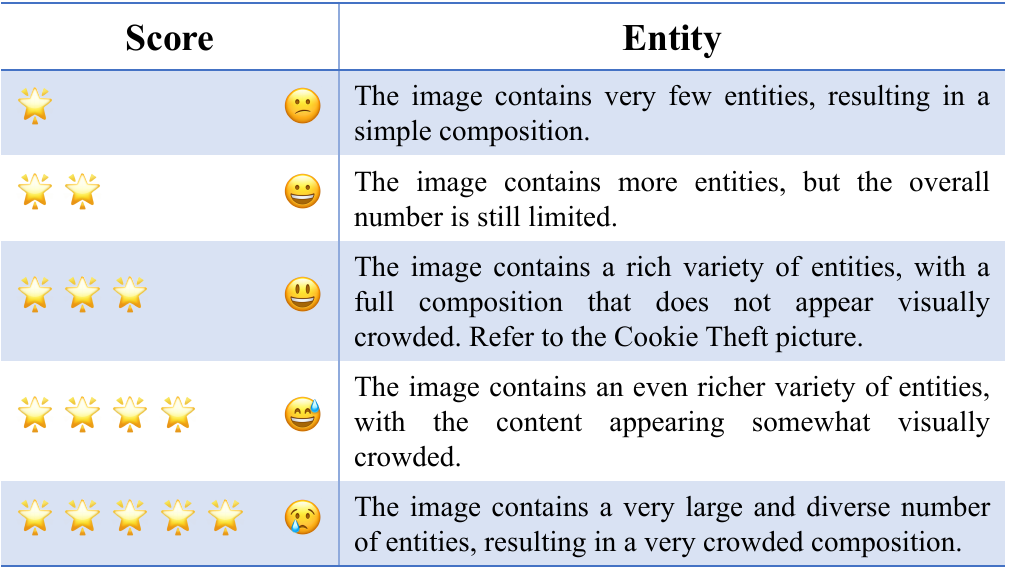}
    \caption{Annotation criteria of Entity Score. The referenced Cookie Theft refers to the updated version.
    }
    \label{entity_ann}
\end{figure}

The scoring criteria (Figure~\ref{entity_ann}) for the Entity Score are based on the richness of entities in the image.
Entity Score differs slightly from that of the ICA task~\cite{ic9600}: we emphasize the richness of entities, e.g., an image with only complex lines would not receive a high score.
Note that, since we do not always expect images to be overly cluttered and overwhelming, a higher Entity Score does not necessarily indicate a better image.

\subsubsection{Semantic Score}

\begin{figure*}[htbp]
    \centering
    \includegraphics[scale=0.55]{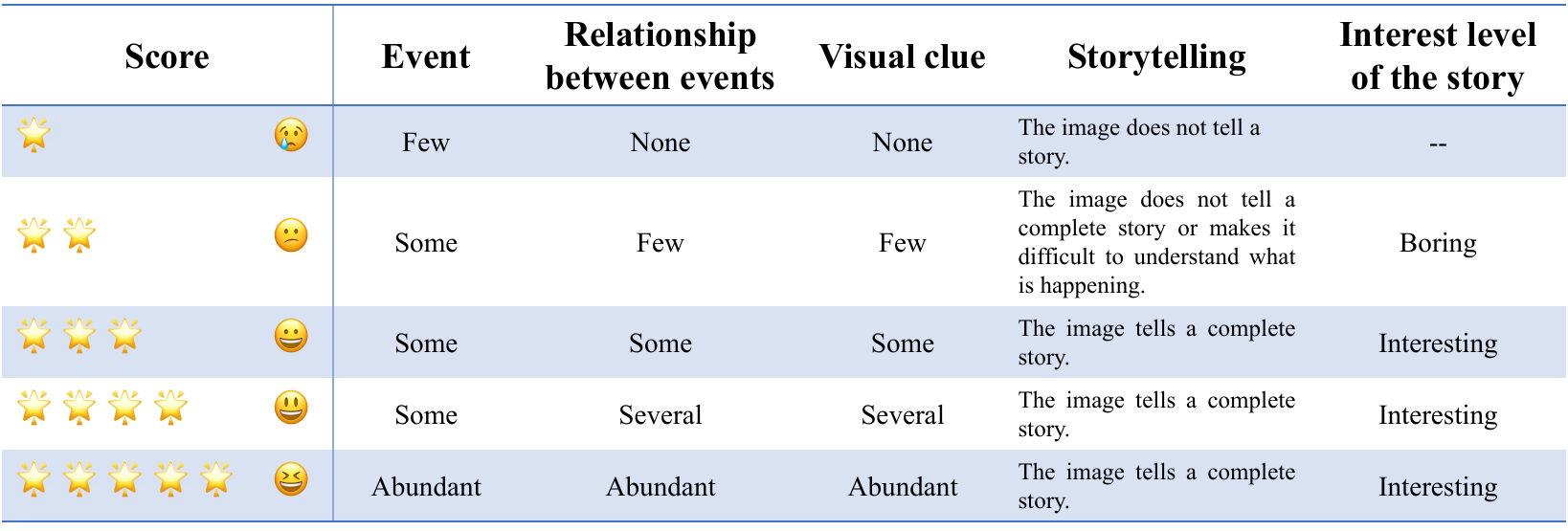}
    \caption{Annotation criteria of Semantic Score.}
    \label{semantic_ann}
\end{figure*}

To help annotators understand semantic complexity, we construct detailed annotation criteria of Semantic Score (Figure~\ref{semantic_ann}).
The scoring criteria consist of five dimensions: event, connection between events, visual clues, storytelling, and interest level of the story.
The connection between events primarily refers to their causal relationships.
For instance, because ``there's a broken bowl on the floor,''  ($\rightarrow$) ``the mother is spanking the boy.''
The definition of visual clues is entities capable of inferring new semantic conclusions. 
Depending on the type of conclusions inferred, visual clues encompass different categories: time, location, characters, character relationships, events, event relationships, character mental states, etc~\cite{song2024cognitive}.
For example, a ``Christmas tree'' suggests the Christmas season, serving as a clue for time reasoning.

Specifically, the interpretation of Figure~\ref{semantic_ann} is as follows:
(1). If the image does not depict any event, or if it features only a very limited variety of events (e.g., running, swimming, etc.), assign it a score of 1.
(2). If there are some kinds of events in the image, but the events are unrelated and there are almost no clues to infer additional information, give it a score of 2.
(3). Assign a score of 2 when there are some events in the image with a slight connection between them or a few clues that suggest additional information, but the image does not convey a clear story or the story's appeal is minimal.
(4). For images rated 3 or above, there must be connections between events and visual clues present in the image. 
(5). The differences between ratings of 3, 4, and 5 primarily lie in the richness and number of these connections and clues.

\subsubsection{Annotation Process}
Each image is annotated by three annotators, each assigning the two scores.
The annotators are mostly undergraduate or graduate students aged 20 to 25.
We begin by providing training to the annotators.
For each annotation score, several examples are provided for them to refer to. 
They are then asked to annotate a small sample set of images as a test first. 
Only annotators who pass the test are allowed to participate in the subsequent formal annotation.

During the annotation process, to ensure labeling quality, we maintain ongoing communication with the annotators, conduct regular spot checks on annotated image groups, and provide prompt feedback on any inaccurate annotations.
If a sample from a particular group has poor annotation results, we will discard the labels for that group and re-label the group of images.
We also provide immediate assistance if they encounter any issues during the annotation process.

\subsection{Dataset Analysis}

\subsubsection{Annotation Consistency}

In line with established standards~\cite{kong2016photo, ying2020patches, ic9600}, we assess the consistency between annotators by using the Pearson Correlation Coefficient (PCC), Spearman's Rank Correlation Coefficient (SRCC), and Kendall's tau correlation for each pair of annotations.
For Entity Score, the average PCC, SRCC, and Kendall's tau are 0.836, 0.827, and 0.762 respectively.
The average PCC, SRCC, and Kendall's tau of Semantic Score are 0.799, 0.798, and 0.729 respectively.
This demonstrates the consistency of our data annotation.
In addition, following the crowdsourcing assessment studies conducted for IQA, IAA, and ICA~\cite{koniq10k, Siahaan2016, ic9600}, we compute the Intra-class Correlation Coefficient (ICC) for our annotations to measure the inter-rater reliability.
The ICCs of Entity Score and Semantic Score are 0.937 and 0.922 respectively, which shows the reliability and consistency of our annotation.

\subsubsection{Dataset Case Analysis}

Figure~\ref{samples} shows some samples of our dataset. 
We can see that images with more entities are scored with higher Entity Scores, and images with more visual clues and telling more engaging stories are scored with higher Semantic Scores.
We can also see that the relationship between Entity Score and Semantic Score is not entirely positively correlated.
Even though some images contain few entities, for example, Figure~\ref{samples} (e), they can still tell an interesting story.
Images with a variety of entities can also contain little semantic information, for instance, Figure~\ref{samples} (d). 

\subsubsection{Dataset Statistics}

Table~\ref{distribution} shows the distribution of our dataset.
We randomly split the data into a training set, a validation set and a test set in a 6:2:2 ratio.

\setlength{\tabcolsep}{1.5mm}
\begin{table}[h]
    \small
    \centering
    \begin{tabular}{l|ccccc}
    \toprule
    Score & [0, 0.2) & [0.2, 0.4) & [0.4, 0.6) & [0.6, 0.8) & [0.8, 1]\\
    \midrule 
    ES & 476 & 789 & 999 & 294 & 388 \\
    SS & 767 & 826 & 828 & 382 & 143 \\
    \bottomrule
\end{tabular}
\caption{Dataset label distribution. ES and SS stand for Entity Score and Semantic Score respectively.}
\label{distribution}
\end{table}

\section{Method}

In order to lay the foundation for the ISA task, we propose a novel baseline method to perform the task.
Since we expect to assess images from a higher semantic level, and language can usually express semantics more directly than images, we believe hybrid utilization of both visual and language information will be helpful for ISA task. 
Thus, we propose the \textbf{V}ision-\textbf{L}anguage collaborative \textbf{ISA} (\textbf{VLISA}) method.

\begin{figure}[H]
    \centering
    \includegraphics[scale=0.33]{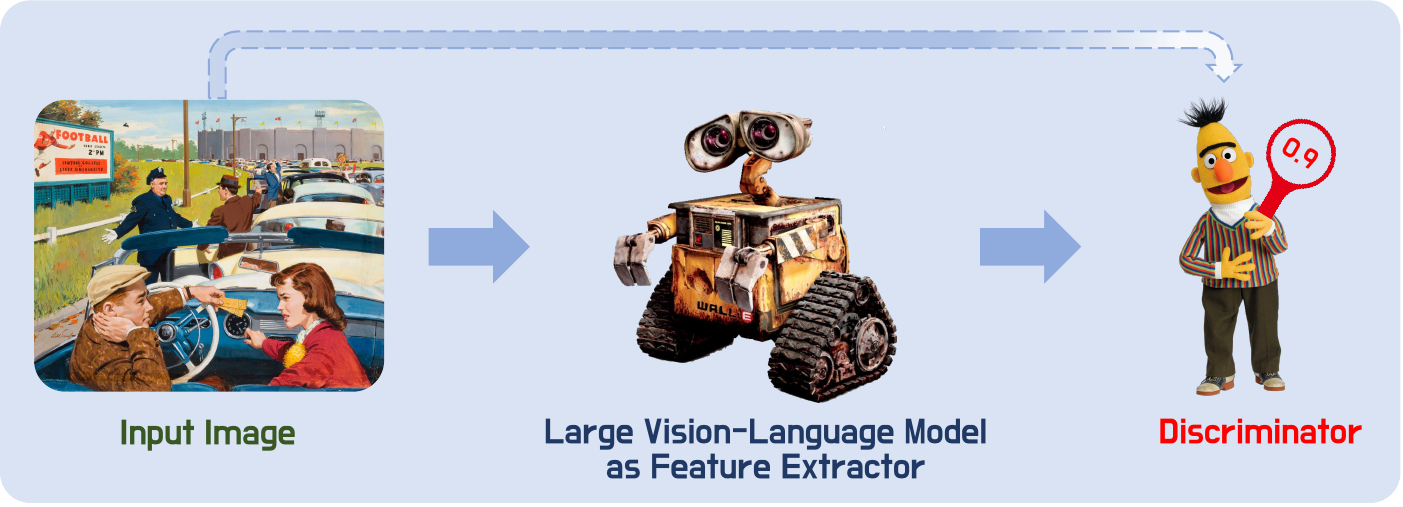}
    \caption{Pipeline of our proposed VLISA method. }
    \label{method}
\end{figure}

As shown in Figure~\ref{method}, VLISA has two components: a Feature Extractor and a Discriminator.
This design follows the typical flow of IAA systems, which consists of a Feature Extraction phase and a Decision Phase~\cite{deng2017image}.
Specifically, we first use an LVLM as the Feature Extractor to extract semantic information in natural language form as features from images.
We adopt GPT-4o~\cite{gpt4} as the default Feature Extractor in this paper, considering its strong capability.
Then, we use a Discriminator model to rate the input image based on the extracted features, optionally including the image itself. 
One advantage of using LVLM as the Feature Extractor is that it can mitigate the impact of different image styles and types and focus more on semantics in images.
Based on different feature extraction modes, we propose two versions of VLISA.

\paragraph{Naive VLISA}

The first way for GPT-4o to extract features from the input image is to make it describe the image in detail.
The prompt is simple: 
\texttt{Describe this image in detail.} 
The generated description is later taken as the input of the Discriminator model.

\paragraph{Chain-of-Thought VLISA} 

The second feature extraction method is inspired by Chain-of-Thought (CoT)~\cite{cot2024}. 
Referring to the annotation instruction of CogBench~\cite{song2024cognitive}, we first ask GPT-4o to generate Chain-of-Reasonings (CoRs) from different aspects, then generate the description based on the CoRs. 
A CoR consists of visual clues and the conclusion drawn from the clues.

Specifically, we adopt seven categories of CoRs:  

\begin{itemize}
    \item \textbf{Special Time.} ``Special time'' refers to a time that requires observation of clues to deduce, rather than an obvious time like ``daytime.''
    \item \textbf{Special Location.} ``Special location'' refers to a location that requires observation of clues to deduce, rather than an obvious location like ``on the roadside.''
    \item \textbf{Character Role.} The roles or identities of characters in the images.
    \item \textbf{Character Relationship.} The relationships between characters in the images.
    \item \textbf{High-level Event.} ``High-level events'' refer to events that require observation of clues to deduce, rather than obvious actions like ``running.''
    \item \textbf{Event Causal Relationship.} The causal relationships between events in the images.
    \item \textbf{Mental State.} Mental states of characters in the images.
\end{itemize}

Both the generated CoRs and description are later used as input for the Discriminator model.

\section{Experiments}

In this section, we present the experimental setup, results, and corresponding analysis.

\begin{table*}[h]
    \centering
    \small
    \begin{tabular}{l|cccc}
        \toprule
        Model & RMSE $\downarrow$ &  RMAE $\downarrow$ &  PCC $\uparrow$ &  SRCC $\uparrow$ \\
        \midrule 
        ICNet & 0.102 (0.001) & 0.281 (0.0003) & 0.918 (0.002) & 0.909 (0.002) \\  
        ViT & 0.094 (0.002) & 0.271 (0.002) & 0.929 (0.002) & 0.923 (0.003) \\ 
        Naive VLISA (ViLT) & \textbf{0.079} (0.002)  & \textbf{0.249} (0.002) & \textbf{0.952} (0.002) & \textbf{0.949} (0.002) \\
        Naive VLISA (BERT) & 0.095 (0.002)  & 0.269 (0.002) & 0.928 (0.001) & 0.925 (0.002) \\ 			
        Naive VLISA (Longformer) & 0.094 (0.0005) & 0.268 (0.0005) & 0.931 (0.0005) & 0.928 (0.0005)  \\			
        CoT VLISA (ViLT) & 0.080 (0.002) & \textbf{0.249} (0.002)  & 0.951 (0.002)  & 0.947 (0.003) \\ 
        CoT VLISA (BERT) &  0.111 (0.001) & 0.288 (0.003)  & 0.902 (0.003)  & 0.897 (0.003)   \\ 
        CoT VLISA (Longformer) & 0.111 (0.001) & 0.287 (0.001) & 0.901 (0.002) & 0.893 (0.001)  \\ 
        \bottomrule
\end{tabular}
\caption{Performance of different methods on the Entity Complexity Scoring task.
$\uparrow$ indicates that the larger the value, the better.
$\downarrow$ indicates that the smaller the value, the better.
}
\label{entity_reg}
\end{table*}

\begin{table*}[h]
    \centering
    \small
    \begin{tabular}{l|cccc}
    \toprule
    Model & RMSE  $\downarrow$ &  RMAE  $\downarrow$ &  PCC $\uparrow$ &  SRCC $\uparrow$ \\
    \midrule 
    ICNet &  0.191 (0.001) & 0.389 (0.002) & 0.634 (0.005) & 0.626 (0.005)  \\  
    ViT & 0.182 (0.001) & 0.383 (0.001) & 0.677 (0.003) & 0.667 (0.004) \\
    Naive VLISA (ViLT) & 0.148 (0.001) & 0.339 (0.002) & 0.799 (0.004)  &  0.791 (0.005) \\ 
    Naive VLISA (BERT) & 0.149 (0.001) & 0.337 (0.002) & 0.798 (0.003) & 0.785 (0.003)  \\ 
    Naive VLISA (Longformer) & 0.148 (0.001) & 0.332 (0.001)  & 0.800 (0.003)  & 0.789 (0.004)  \\
    CoT VLISA (ViLT) & \textbf{0.140} (0.0005) & \textbf{0.328} (0.001) & \textbf{0.823} (0.002) & \textbf{0.817} (0.003)  \\ 
    CoT VLISA (BERT) & 0.144 (0.0005) & \textbf{0.328} (0.001) & 0.812 (0.001) & 0.802 (0.002)  \\ 
    CoT VLISA (Longformer) & 0.149 (0.002) & 0.333 (0.002) & 0.804 (0.005) & 0.795 (0.007)  \\
    \bottomrule
\end{tabular}
\caption{Performance of different methods on the Semantic Complexity Scoring task.}
\label{semantic_reg}
\end{table*}

\begin{table*}[h]
    \centering
    \small
    \begin{tabular}{l|c|cccc}
    \toprule
    Task & Model & RMSE $\downarrow$ &  RMAE $\downarrow$ &  PCC $\uparrow$ &  SRCC $\uparrow$ \\
    \midrule 
           & Naive VLISA (ViLT) & 0.080 (0.000) & 0.247 (0.0005) & 0.951 (0.0005) & 0.948 (0.0005)  \\
    Entity & Naive VLISA (BERT) & 0.096 (0.001) & 0.268 (0.001) & 0.927 (0.001) & 0.929 (0.0005)  \\
           & Naive VLISA (Longformer) & 0.100 (0.003) & 0.276 (0.004) & 0.924 (0.001) & 0.924 (0.001) \\
    \hline
             & Naive VLISA (ViLT) & 0.149 (0.001) & 0.343 (0.0005) & 0.799 (0.001) & 0.789 (0.001)  \\
    Semantic & Naive VLISA (BERT) & 0.158 (0.002) & 0.349 (0.001) & 0.770 (0.006) & 0.757 (0.004)  \\
             & Naive VLISA (Longformer) & 0.152 (0.001) & 0.339 (0.0005) & 0.790 (0.003) & 0.780 (0.001) \\
    \bottomrule
\end{tabular}
\caption{Performance of Naive VLISA with CogVLM2 as the Feature Extractor. Entity and Semantic
        refer to the Entity Complexity Scoring task and the Semantic Complexity Scoring task, respectively.
}
\label{cogvlm}
\end{table*}

\begin{table*}[h]
    \centering
    \small
    \begin{tabular}{l|c|cccc}
    \toprule
    Model & CoT Feature & RMSE $\downarrow$ &  RMAE $\downarrow$ &  PCC $\uparrow$ &  SRCC $\uparrow$ \\
    \midrule 
                     & CoRs + Description & 0.140 (0.0005) & 0.328 (0.001) & 0.823 (0.002) & 0.817 (0.003)  \\
    CoT VLISA (ViLT) & w/o CoRs & 0.143 (0.001) & 0.335 (0.001) & 0.818 (0.003) & 0.808 (0.003)  \\
                     & w/o Description & 0.142 (0.002) & 0.330 (0.004) & 0.819 (0.007) & 0.808 (0.008) \\
    \hline
                     & CoRs + Description & 0.144 (0.0005) & 0.328 (0.001) & 0.812 (0.001) & 0.802 (0.002)  \\
    CoT VLISA (BERT) & w/o CoRs & 0.145 (0.000) & 0.332 (0.0005) & 0.809 (0.001) & 0.799 (0.001) \\
                     & w/o Description & 0.148 (0.001) & 0.332 (0.001) & 0.801 (0.004) & 0.792 (0.005) \\
    \hline
                           & CoRs + Description & 0.149 (0.002) & 0.333 (0.002) & 0.804 (0.005) & 0.795 (0.007) \\
    CoT VLISA (Longformer) & w/o CoRs & 0.148 (0.001) & 0.334 (0.002) & 0.802 (0.003) & 0.793 (0.001) \\
                           & w/o Description & 0.149 (0.001) & 0.331 (0.001) & 0.800 (0.004) & 0.791 (0.005) \\
    \bottomrule
\end{tabular}
\caption{Ablation study of CoT VLISA on the Semantic Complexity Scoring task.}
\label{ablation}
\end{table*}

\paragraph{Models}
For vision models, we use ICNet~\cite{ic9600} and ViT~\cite{vit} as our baseline models. 
ICNet is a model designed for the ICA task. 
ViT is a classic vision model. 
For VLISA, we use GPT-4o~\cite{gpt4} to extract features from the image and use ViLT~\cite{Kim2021ViLTVT}, BERT~\cite{devlin-etal-2019-bert}, and Longformer~\cite{Beltagy2020Longformer} as the Discriminator models. 
ViLT accepts both an image and its text features as input, while BERT and Longformer only accept text features as input.

\paragraph{Implementation Details}
We implement the models using PyTorch~\cite{torch} and Transformers~\cite{wolf-etal-2020-transformers}. 
Each model is trained and evaluated on either a single NVIDIA A10 GPU or a Tesla V100 GPU.
For ICNet, we mainly follow the settings in the original paper.
For ViT, ViLT, BERT, and Longformer, we train them with batch size 16. 
They are fine-tuned based on \textit{vit-base-patch16-224}, \textit{vilt-b32-mlm}, \textit{bert-base-uncased}, and \textit{longformer-base-4096}, respectively.
The maximum text input length of ViLT and BERT is set to 512 tokens.
Since the maximum text input length of pre-trained ViLT is 40, we randomly initialize its position embeddings.
The maximum input length of Longformer is set to 1024 tokens.
These models are trained, validated, and tested on our training, validation, and test sets, respectively.
We repeat all experiments three times and calculate the mean and standard deviation.

\paragraph{Evaluation Metrics}

Following~\citet{ic9600}, we use Root Mean Square Error (RMSE), Root Mean Absolute Error (RMAE), Pearson Correlation Coefficient (PCC), and Spearman's Rank Correlation Coefficient (SRCC) as our evaluation metrics.
The formulas for calculating RMSE and RMAE are

\begin{equation}
RMSE = \sqrt{ \frac{1}{n} \sum\nolimits_{i=1}^{n} (x_i - y_i)^2 }
\end{equation}

and

\begin{equation}
RMAE = \sqrt{ \frac{1}{n} \sum\nolimits_{i=1}^{n} | x_i - y_i |},
\end{equation}
where $n$ is the sample size,  $x_i$ and $y_i$ represent the $i$th label and  predicted score. 

The PCC is defined as

\begin{equation}
r_{p} = \frac{\displaystyle\sum\nolimits_{i=1}^{n}(x_i-\overline{x})(y_i-\overline{y})}{\sqrt{\displaystyle\sum\nolimits_{i=1}^{n}(x_i-\overline{x})^2}\sqrt{\displaystyle\sum\nolimits_{i=1}^{n}(y_i-\overline{y})^2}},
\end{equation}
where $\overline{x}$ and $\overline{y}$ are the means of $\mathbf{x}$ and $\mathbf{y}$.

The formula of SRCC is

\begin{equation}
r_s = 1 - \frac{6 \sum_{i=1}^{n} d_i^2}{n(n^2 - 1)},
\end{equation}
where $d_i=R(x_i)-R(y_i)$, $R(x_i)$ and $R(y_i)$ are the ranks of the $i$th image when labels and predicted scores are sorted in descending order.

\paragraph{Results}

Table~\ref{entity_reg} shows the results of the Entity Complexity Scoring task.
Naive VLISA with a pre-trained language model (BERT/Longformer) as the Discriminator shows competitive performance compared to ViT.
When both images and text features are input to the ViLT Discriminator, the model performs significantly better than ViT and other Naive VLISAs.
Naive VLISA (BERT/Longformer) outperforms CoT VLISA (BERT/Longformer).
One possible reason is that features extracted by Naive VLISA tend to focus more on describing the content at the entity level within the image.
Conversely, the Feature Extractor in CoT VLISA extracts higher-level semantic information from the image, but it overlooks some entities.
Naive VLISA (ViLT) is less affected by the type of text features, probably because the image itself is visible to it.

Table~\ref{semantic_reg} shows the results of the Semantic Complexity Scoring task.
We can see that predicting Semantic Score is more challenging than predicting Entity Score and the traditional vision models cannot perform well on this task.
Naive VLISAs show obviously better performance than ViT and ICNet.
The possible reason is that with GPT-4o extracting semantic information from the images, the Discriminator in VLISA can perform score prediction at a higher semantic level.
Consistent with the previous hypothesis, CoT VLISA shows better performance than Naive VLISA on this task.
CoT VLISA (ViLT) shows the best performance.
Comparing the performance of VLISAs and vision models on the two tasks highlights the importance of introducing the language modality for the Semantic Complexity Scoring task.

Generally speaking, Naive VLISA can perform well on both sub-tasks, and CoT VLISA can further improve the performance on the Semantic Complexity Scoring task.

\paragraph{Open-source LVLM as Feature Extractor}

To further validate the effectiveness of VLISA pipeline, we replace GPT-4o with an open-source LVLM, CogVLM2~\cite{hong2024cogvlm2}, as the Feature Extractor in Naive VLISA. 
As shown in Table~\ref{cogvlm}, we observe that using CogVLM2 as the Feature Extractor does not significantly degrade the models' performance, especially for Naive VLISA (ViLT). 
This demonstrates the robustness of our approach.

\paragraph{Ablation Study}

CoRs and description are the two main parts extracted by the Feature Extractor of CoT VLISA, so we validate their effectiveness in this section.
Table~\ref{ablation} shows when either CoRs or description are removed from the extracted text features, there may be a slight performance drop. Therefore, we recommend using both the CoRs and description as input.

\section{Analysis}

\begin{figure*}[ht]
    \centering
    \includegraphics[scale=0.46]{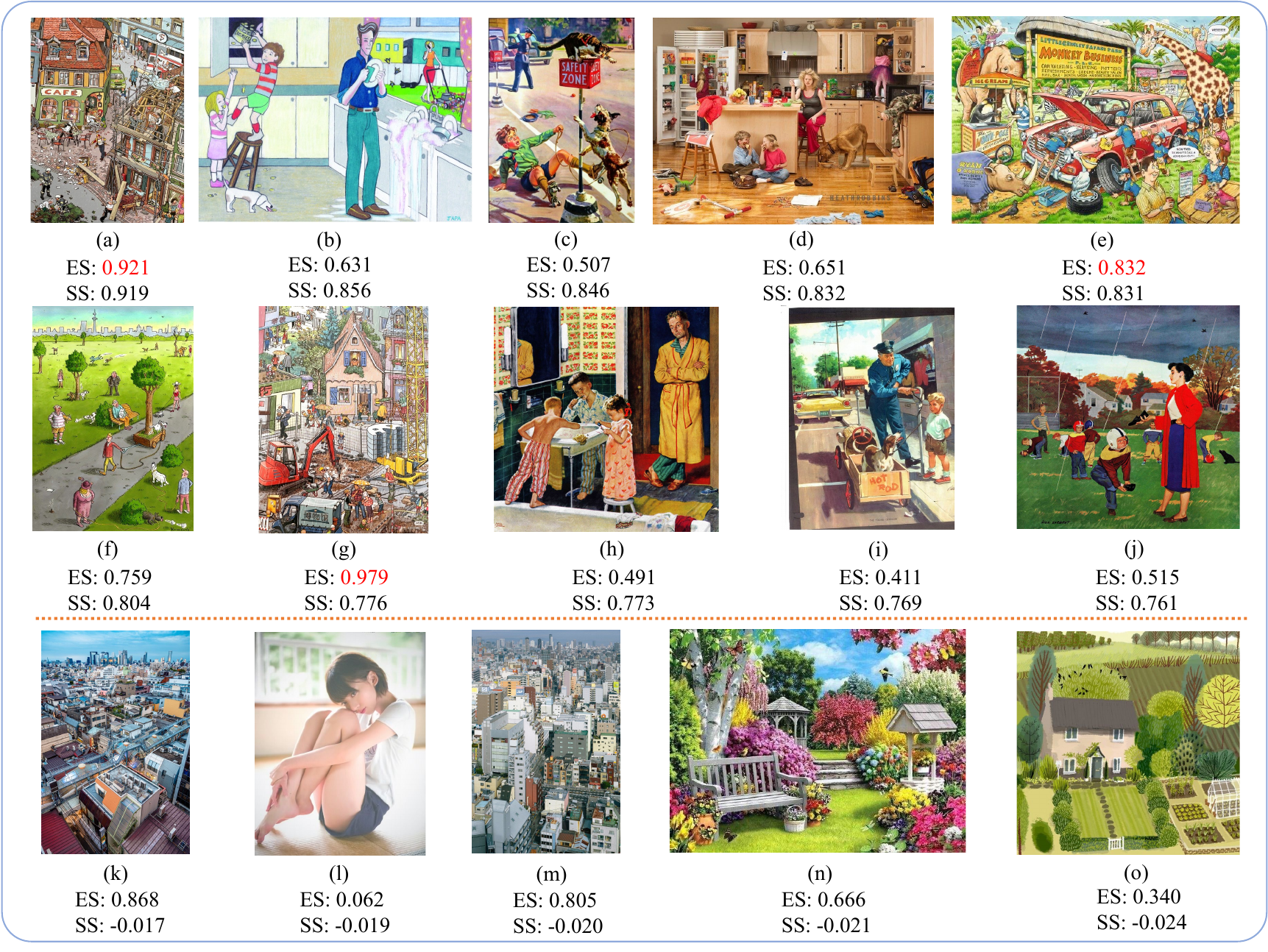}
    \caption{Case study. ES and SS stand for Entity Score and Semantic Score predicted by Naive VLISA (Longformer) respectively. The samples above the orange line are those with the highest Semantic Scores, sorted in descending order. The samples below the line are those with the lowest Semantic Scores. Red Entity Scores indicate that they are too high. 
    }
    \label{pred_case}
\end{figure*}

When using VLISA to identify semantically rich images, we recommend starting with those that have high Semantic Scores and then refining the selection based on the Entity Score to match the application scenario.
Figure~\ref{pred_case} shows some samples with scores predicted by Naive VLISA (Longformer).
In the scenario of searching for images similar to the Cookie Theft picture, images with higher Semantic Scores and moderate Entity Scores are preferred, according to the design guidelines.

In Figure~\ref{pred_case}, we can see that images with the highest Semantic Scores generally tell more compelling stories, containing richer semantic information. 
Interestingly, without including the Cookie Theft image in the training set, the image with the second-highest Semantic Score is Cookie Theft (Figure~\ref{pred_case} (b)).
Based on Entity Score, we can further filter out images with too few or too many entities. 
For instance, (a), (e), and (g) contain too many entities, though they also tell interesting stories. 
That is, the remaining images above the orange line are more preferred by the Cookie Theft design principles.
Note that although many images in our dataset are not real-world images, VLISA can still give appropriate Semantic Scores to these real-world images.
For example, image (d) in Figure~\ref{pred_case} has a high Semantic Score and it is actually quite similar to the Cookie Theft semantically.
This demonstrates that our method can avoid the influence of types or styles of images to some extent.

For images with the lowest Semantic Scores under the orange line, there is either a single action (Figure~\ref{pred_case} (l)) or no event at all (Figure~\ref{pred_case} (k, m, n, o)), which is also consistent with our annotation design.

\section{Related Work}

\paragraph{Image Quality Assessment}
Image Quality Assessment (IQA) is a task to assess the quality of images.
It mainly concerns various types of distortions introduced in stages of the visual communication systems.
The rapid growth of visual media has driven the development of many IQA methods~\cite{zhai2020perceptual}.
Some IQA datasets including TID2013~\cite{tid2013}, KonIQ-10k~\cite{koniq10k}, SPAQ~\cite{Fang_2020_CVPR} and PaQ-2-PiQ~\cite{ying2020patches} etc. are proposed.
With the development of LVLMs, \citet{wu2024qbench} propose Q-Bench to assess their abilities on low-level visual perception and understanding, which plays significant roles in IQA.
The difference between IQA and our ISA task is that ISA task focuses on analyzing the semantic content~\cite{zhang2015association} of an image rather than its quality.

\paragraph{Image Aesthetics Assessment}
Different from IQA, Image Aesthetics Assessment (IAA) task assesses the aesthetics of an image from the perspective of its content.
Typical IAA task seeks to computationally assess the quality of photos based on photographic rules~\cite{deng2017image}.
Several IAA datasets are proposed, for example, the Photo.net dataset~\cite{2011Photonet}, the DPChallenge dataset~\cite{2008DPChallenge}, and the TAD66K dataset~\cite{ijcai2022p132} etc.
As the development of image style transfer and AI painting, Artistic Image Aesthetic Assessment (AIAA) task is proposed to automatically evaluate artwork aesthetics~\cite{2015JenAesthetics, fekete2022vienna, Yi_2023_CVPR}.
The difference between IAA and ISA task is that ISA assesses images based on their semantic richness.

\paragraph{Image Complexity Assessment}

Image Complexity Assessment (ICA) is proposed to assess the intricacy contained within an image~\cite{visc2009}.
It measures the richness of details and diversity within the image~\cite{Snodgrass1980ASS}.
The SAVOIAS dataset~\cite{saraee2020visual} contains over 1,000 images and labels for IC analysis.
\citet{ic9600} built the first large-scale IC dataset with 9,600 annotated images IC9600 dataset and proposed a baseline model called ICNet.
Compared to IQA and IAA, ICA task is more relevant to ISA task. 
Despite the differences, the Entity Richness Scoring in ISA shares similarities with the ICA task.
However, the key distinction lies in ISA’s emphasis on a higher semantic level~\cite{li2015large}, rather than merely evaluating complexity at the entity level.

\section{Conclusion}

In this paper, we propose a novel ISA task to identify storytelling images with rich semantics.
We propose the first ISA dataset consisting of an Entity Complexity Scoring task and a Semantic Complexity Scoring task.
We also propose a simple yet effective method called VLISA as our baseline model for this task.
We believe this task will have a wide range of applications in the future.
For example, with the Entity Score and Semantic Score, images with different semantic complexity levels can be selected.
It can also facilitate AI models in understanding and generating images with richer semantics.

\appendix

\section*{Ethical Statement}

We follow the Terms of Service of Pinterest to collect the images in our ISA dataset.
Our dataset will be available to download for research purposes only, which is in compliance with the Terms of Service of Pinterest.

\section*{Acknowledgments}
Mengyue Wu was supported by National Natural Science Foundation of China (Grant No.92048205), the CMB Credit Card Center-SJTU joint research grant, Shanghai Municipal Science and Technology Major Project (2021SHZDZX0102), Jiangsu Technology Project (No.BE2022059-2) and Guangxi Major Science and Technology Project~(No. AA23062062). 
Kenny Q. Zhu was partially supported by National Science Foundation (NSF) Award (No. 2349713).

\bibliography{aaai25}

\end{document}